# A new semantically annotated corpus with syntactic-semantic and cross-lingual senses

**Myriam Rakho[a], Éric Laporte[b] and Matthieu Constant[c]**
[a,b,c]Université Paris-Est, LIGM
E-mail: {rakho,eric.laporte,Matthieu.Constant}@univ-mlv.fr

**Abstract**

In this article, we describe a new sense-tagged corpus for Word Sense Disambiguation. The corpus is constituted of instances of 20 French polysemous verbs. Each verb instance is annotated with three sense labels: (1) the actual translation of the verb in the english version of this instance in a parallel corpus, (2) an entry of the verb in a computational dictionary of French (the Lexicon-Grammar tables) and (3) a fine-grained sense label resulting from the concatenation of the translation and the Lexicon-Grammar entry.

## 1 Introduction

Word Sense Disambiguation (WSD) is the task of automatically identifying the correct sense of a polysemous word in particular contexts. WSD is an intermediate NLP task, and thus some key factors in designing the disambiguation strategy depend on the final application, among them the sense inventory definition. One common practice in recent multilingual WSD is to use the actual translations of the polysemous words in a parallel corpus as sense labels. Another approach, less experimented, has appeared recently which uses the entries of the polysemous words in a computational lexicon as sense labels. The level of sense granularity in those two approaches may not be sufficient for multilingual WSD when they are used individually. In both cases, some sense labels may be to coarse-grained or too fine-grained for the source language (SL) or for the target language (TL).

In this paper, we address the problem of identifying an appropriate and effective WSD sense inventory for machine translation (MT) applications. As suggested by Ide and Wilks (2006), we propose to combine evidence from the translations and from the entries in a computational lexicon to determine the polysemous word senses. For that purpose, we have built a new semantically annotated corpus for 20 French polysemous verbs (section 2). In this corpus, every verb instance is tagged with 4 sense labels:

- **(1)** its corresponding entry in a computational lexicon in which sense distinctions are based on syntactic-semantic evidence, the Lexicon-Grammar tables (denoted as LG), a computational dictionary for French in which the senses of the words are determined on the basis of syntactic and semantic evidence
- **(2)** its actual translation in a parallel corpus (TRSL)
- **(3)** a fine-grained sense label which is concatenation of the LG entry and the translation (LG#TRSL)
- **(4)** and a class of sense labels: all the fine-grained senses that have the same 'LG' part (i.e. all the possible translations of its LG sense tag)

### 1.1 Translational sense inventories

TL oriented sense inventories for multilingual WSD are usually built-up from parallel multilingual corpora. The translations of the polysemous words are obtained by automatic word alignment of the source and target versions of a corpus. Word senses are either single translations or clusters of translations. The best example of using such a sense inventory is the Cross-Lingual WSD task in the SemEval-2 (Lefever and Hoste, 2009) and in the forthcoming SemEval-3 exercises. The test datasets were then manually annotated by native speakers with manually built clusters of TL translations.

**Advantages.** From the theoretical point of view, this method makes sense both in terms of motivations and in terms of ressources' availability. On one hand, the senses are well-motivated : they are completely application-oriented, the senses are differentiated only when they are lexicalized with different words in the TL. And since the sense inventory is extracted from the corpus, the senses are attested in the datasets (Edmonds and Kilgariff, 2002). On the other hand, there is no need for predefined sense inventories and the annotated corpora can be built automatically, which is one great benefit for this approach.

**Drawback.** From the computational point of view, the drawback of this approach arises from the fact that the sense are differentiated on the basis of cross-linguistic evidence whilst the task of WSD itself consists in identifying the correct sense of a word by using 'distributional' evidence from its context, typically syntactic and semantic information. Cross-lingual sense distinctions do not necessarily correspond with distributional sense distinctions. So assigning the same cross-lingual sense label to contexts with different distributional characteristics (contexts that represent different SL senses) affects the homogeneity of the description of this sense in terms of linguistic features. This leads to an increase in the task difficulty.

For example, two senses of the French verb *comprendre* are used in *Nous* ***comprenons*** *votre inquiétude* and in *Les personnes qui commettent des fraudes doivent* ***comprendre*** *qu'elles seront poursuivies*. Both occurrences of this verb can be translated to English as *understand* : *We* ***understand*** *your concerns* and *People need to* ***understand*** *that if they commit fraud they will be prosecuted* (respectively). So, besides the translation, a WSD system would need more information in order to annotate correctly those two occurrences.

### 1.2 Lexicon-style sense inventories

A novel approach to sense inventory definition for WSD have been proposed by Abend et al. (2007), Chen and Di Eugenio (2010) and Brown *et al.* (2011). They experimented WSD as the task of classifying verb instances into their entries in VerbNet (Kippler-Schuler *et al.*, 2007), the largest verb lexicon currently available for English. VN extends Levin's (1993) verb classification both in coverage and in the verbs' description. It is the most similar ressource to the LG tables. Those experiments reported results that overcome the state-of-the-art of WSD with cross-lingual sense distinctions.

**Advantages.** The senses here are characterized exclusively by means of the syntactic and semantic characteristics of the SL contexts in which they are use. And since the task of WSD is to identify the correct meaning by using typically this kind of information about the context, this approach is more convenient, from the computational point of view.

For example, in the two sentences of the example of section 2.1, the two senses would be described by 2 SCFs : when used in the first sense, the verb selects a direct object (*votre inquiétude*) or a subjonctive clause complement whilst it selects an indicative clause complement (*qu'elles seront poursuivies*) in the second sense. Modeling such kinds of information in a feature space representation that would be used to train a WSD system can be done by using a syntactic parser.

**Drawback.** Sense distinctions based on contextual information in the SL may be too coarse-grained. Gale *et al.* (1993) illustrate this kind on semantic ambiguity with the English verb 'wear' which is translated to Japanese with 5 different words depending of what part of the body is involved.

## 2 A new semantically annotated corpus for WSD

As a remedy to the deficiences of the cross-lingual and the lexicon-style sense labels, we have constructed a new corpus for 20 French polysemous verbs. First, we have built 4 different sense inventories for every verb : the list of its actual translations in a parallel corpus, its entries in the Lexicon-Grammar tables, and two sense inventories automatically generated by combining the two firsts. Then each verb instance have been assigned 4 sense labels, one from every sense inventory.

In the 2 next subsections, we describe the 20 polysemous verbs and the construction of their corpora. The 4 following subsections describe our 4 sense inventories and our method for data annotation.

### 2.1 Corpus construction

We have built a corpus for the 20 French polysemous verbs listed in the table 1 below. Those verbs have been selected on the basis of their polysemy during the ARCADE campaign (2000) for the evaluation of multilingual word alignment systems.

| *arrêter, comprendre, conclure, conduire, connaître, couvrir, entrer, exercer, importer, mettre, ouvrir, parvenir, passer, porter, poursuivre, présenter, rendre, répondre, tirer, venir* |
|---|

Table 1: The test French polysemous verbs

For the construction of the corpus, we have used the EuroParl-Intersection corpus (Koehn, 2005), a multilingual corpus in which every sentence is aligned with its translations in 6 different languages. We have aligned the French (our SL) and the English (our TL) versions of this corpus at the word level with the GIZA++ tool (Och and Ney, 2003). And then, from the French version, we have extracted all the contexts in which one of the 20 polysemous verbs occurs along with their translation in the English version.

With GIZA++, a single token can be aligned with a sequence of one or more TL tokens but the inverse is not allowed (i.e. aligning a sequence or SL tokens with a TL token). So we have performed two word alignments : one from SL to TL (*fr-en*) and one from TL to SL (*en-fr*). And we have eliminated all the couples of contexts in which the TL word aligned with the polysemous verb in the *en-fr* alignment did not appear in the words aligned with it in the *fr-en* alignment. For example, the following sentence is a context of the verb *comprendre* :

**(2)** **Fr:** *Les personnes commettant des fraudes doivent* ***comprendre*** *qu'elles seront poursuivies.*

**En:** *People need to* ***understand*** *that if they commit fraud they will be prosecuted.*

The *fr-en* and *en-fr* alignments have provided, respectively, the couples (*comprendre – understand*) and (*understand – comprendre*). So *understand* has been considered as a possible translation of *comprendre*. In another context of *comprendre*, the two alignments have provided the couples (*comprendre – find*) and (*incomprehensible – comprendre pas*). So *find* has been considered as a non valid translation of *comprendre*.

After that, the translations have been validated manually.

## 2.2 The translational sense inventory

**Sense inventory.** We have built for every polysemous verb a translational sense inventory : its TL translations obtained from the word alignment of its SL and TL contexts.

**Data annotation.** Annotating the corpora with this sense inventory was done automatically and validated manually. For example, the following occurrence of the verb *comprendre* has been assigned the label *understand* :

**(3)** **Fr:** *C'est inadmissible et je **comprends** l'impatience des personnes qui exigent le remplacement de ces instruments par un procureur européen commun*.

**En:** *It is simply too bad, and I can well **understand** why people are becoming impatient and demanding that we have a common European Public Prosecutor's Office instead.*

## 2.3. The lexicon-style sense inventory

**Sense inventory.** We have also built a lexicon-style sense inventory by using the Lexicon-Grammar tables, which is currently one of the major sources of syntactic lexical information for the French language.

The Lexicon-Grammar is organised into a series of tables, each of them grouping lexical items which share a certain number of *defining features*. The verbs belonging to the same table form a syntactically and semantically coherent class. The underlying assumption in the Lexicon-Grammar (LG) is that entry distinctions correspond to sense distinctions. The verb entries are described in terms of syntactic and semantic features. Each verb has as many entries in the LG tables (sometimes in the same table) as it has different possible senses. For the data annotation, we have denoted a verb entry as {*V*,*C*}*_table_entry* where the *C* (or *V*) letter means that the verb is in a table that contains idiomatic expressions (or not, resp.), *table* is the name of a table and *entry* is the unique identifier of the verb in this table.

For example, the French verb *comprendre* [understand, comprise] has 9 entries in 9 different tables. The 9 examples above illustrate them.

**(4)** V_6_73

*Max a **compris** qu'Ida était coupable à cet indice*

Max has understood that Ida was guilty from this indication

**(5)** V_10-46

*Faire ce travail **comprend** pour Max qu'il nettoie tout*

To make this work includes for Max that he cleans everything

**(6)** V_12_17

*Max **comprend** qu'Ida ne vienne pas*

Max understands that Ida does not come

**(7)** V32NM_26

*Ce livre **comprend** dix chapitres*

This book comprises ten chapters

**(8)** V_32R3_178

*Max **comprend** (cette langue + l'anglais)*

Max understands this language

**(9)** V_38LR_48

*Le garçon **comprend** le service dans la note*

The boy includes the service in the note

**(10)** V_38R_54

*Max a **compris** cette remarque comme une plaisanterie*

Max has understood this remark as a joke

**(11)** V_31H_124

*Max se **comprend** quand il dit cela*

Max knows what he is trying to say

**(12)** C_c7_49

*$ne **comprendre**$ rien*

to have no clue

**Data annotation.** The data annotation with this sense inventory was manual. The example above (section 2.2) was labeled *V_12_17* : a non idiomatic expression, the 17[th] entry in table 12.

## 2.4 The fine-grained sense inventory

**Sense inventory.** We have generated two new sense inventories by merging automatically the translations and the LG entries. For every context in the corpus we have concatenated its LG entry and its translation labels in the form of *LG#translation.*

**Data annotation.** The example of section 2.2 have been labeled as *V_12_17#understand*.

## 2.5 The clustered fine-grained sense inventory

**Sense inventory.** The labels from the third sense inventory have also been clustered automatically by LG entry : all sense labels involving the same LG entry (i.e. all its possible translations) belong to the same cluster.

**Data annotation.** The contexts have been automatically assigned the same cluster as their label from the third sense inventory. The example of section 2.2 have been assigned the following cluster :{*V_12_17#understand*; *V_12_17#appreciate*; *V_12_17#sympathise*; *V_12_17#realise*; *V_12_17#understandable*; *V_12_17#incomprehensible*; *V_12_17#sympathise*; *V_12_17#sympathy*; *V_12_17#comprehend*; *V_12_17#accept*; *V_12_17#know*; *V_12_17#recognise*; *V_12_17#grasp*; *V_12_17#acknowledge*; *V_12_17#aware*; etc.}.

## 2.6 The clustered fine-grained sense inventory

Table 1 is a statistic description of the sense tagged corpus: the sample size and the number of senses in each sense inventory for each word.

| Target words | Num samples | Num LG | Num TRSL | Num LG#TRSL |
|---|---|---|---|---|
| *arrêter* | 2033 | 12 | 150 | 242 |
| *comprendre* | **8240** | 8 | 183 | 308 |
| *conclure* | 3488 | 5 | 79 | 122 |
| *conduire* | 2114 | 10 | 96 | 145 |
| *connaître* | 5786 | 14 | 158 | 238 |
| *couvrir* | 2183 | 16 | 85 | 128 |
| *entrer* | 2325 | 6 | 107 | 189 |
| *exercer* | **1851** | 4 | 86 | 120 |
| *importer* | 2778 | 5 | **71** | **101** |
| *ouvrir* | 2656 | 17 | 127 | 253 |
| *parvenir* | 7469 | **3** | 152 | 185 |
| *porter* | 3301 | **20** | **219** | **495** |
| *poursuivre* | 5354 | 5 | 154 | 219 |
| *rendre* | 6731 | 14 | 177 | 347 |
| *tirer* | 2163 | 19 | 102 | 160 |
| *venir* | 7369 | 12 | 120 | 306 |
| **Overall mean** | **3837** | **11** | **129** | **222** |

Table 2: Target words samples size and number of senses in each of their 3 first sense inventories

## 3 Conclusion

We have built a new sense tagged corpus with 4 sense inventories based on different kinds of evidence. First, we have annotated our corpus with a sense inventory in which sense distinctions are based on cross-lingual evidence: the TL translations. The level of granularity in such sense inventory may not be appropriate for NLP applications. As a solution, we have annotated our corpus with a second sense inventory in which sense distinctions are based on syntactic-semantic evidence: the LG entries. This sense inventory also poses a problem of sense granularity. So we have combined those two sense inventories into two new sense inventories that combine cross-lingual and syntactic-semantic evidence for sense distinction. In the 3rd sense inventory, each LG sense is split into as many fine-grained senses as it has possible translations, and each TL translation is split into as many fine-grained senses as it has possible corresponding SL senses (LG entries). Finally, from this 3rd sense inventory, we have built a 4th one by clustering automatically the TL translations that correspond to the same LG entry.

We have annotated a corpus with those 4 sense inventories for 20 French polysemous verbs and we plan to create a similar corpus for 20 French polysemous nouns and 20 adjectives.